\documentclass[11pt,a4paper]{article}
\usepackage[hyperref]{acl2020}
\usepackage{times}
\usepackage{latexsym}
\usepackage{comment}
\usepackage{times}
\usepackage{latexsym}
\usepackage{xcolor}
\usepackage{url}
\usepackage{algorithm2e}
\usepackage{graphicx}
\usepackage{amsmath}
\usepackage{fancyhdr}

\usepackage{microtype}

\aclfinalcopy 

\title{Federated pretraining and fine tuning of BERT using  clinical notes from multiple silos }

\author{Dianbo Liu \\
  Boston Children's Hospital  \\
  Harvard Medical School  \\
   \texttt{dianbo.liu@childrens.harvard.edu}\\\And
   Tim Miller \\
  Boston Children's Hospital  \\
  Harvard Medical School  \\
  \texttt{Timothy.Miller@} \\}

\date{}

\begin{document}

\maketitle
\begin{abstract}
Large scale contextual representation models, such as BERT, have significantly advanced natural language processing (NLP) in recently years.
%
However, in certain area like healthcare, accessing diverse large scale text data from multiple institutions is extremely challenging due to privacy and regulatory reasons. 
In this article, we show that it is possible to both pre-train and fine tune BERT models in a federated manner using clinical texts from different silos without moving the data. 
\end{abstract}

\section{Introduction}


In recent years, natural language processing (NLP) has been revolutionized by large contextual representation models pre-trained with large amount of data such as ELMo \cite{peters2018deep} and BERT \cite{devlin2018bert}.  Compared with traditional word level representations, such as word2vec \cite{mikolov2013distributed}, GloVe \cite{pennington2014glove} and fastText \cite{bojanowski2017enriching}, which assign a representation vector to a word regardless of their surround context, contextual representation method models context of a word. For example, the embedings of the word ``bank'' are different between the context of``Bank of England'' and ``the bank of the Charles river." ELMo~\cite{peters2018deep} uses a recurrent neural network model to learn information of contexts of words from unlabelled texts. The trained contextual embedding were then used for downstream tasks. More recently, another contextual word representation model based on transformer, called BERT, was released, and has become widely used for many NLP tasks  \cite{vaswani2017attention,devlin2018bert}. Instead of conducting directional modeling of context of a word, transformers like BERT model relations between all pairs of tokens using a self-supervised strategy. Advances in these contextual representation based model have created new state-of-the-art performance in many NLP benchmark tasks.

Recent studies have shown that training contextual representations in texts from specific domains improves power of the model by capturing domain specific linguistic characteristics. 
Texts from biomedical publications and electronic medical record have been used to pre-train BERT models for NLP task in this domain and showed considerable improvement in many downstream tasks \cite{lee2019biobert,alsentzer2019publicly,si2019enhancing}.


Despite the success of the efforts mentioned above, many challenges remain. Training a medically useful and generalizable contextual representation model requires access to a large amount of clinical texts. In addition, as the way clinicians write texts vary significantly among hospitals and healthcare systems, it is important to have access to data from many sources. 
However, sharing clinical data is difficult due to privacy and regulatory issues. Training NLP models in a federated manner is a good option to overcome these challenges. In this article, we conduct a proof-of-concept study to train BERT across clinical notes from multiple sites in a federated manner without moving notes outside of their silos. Our main contribution include:
\par
1. We show that it is possible to conduct federated pre-training of BERT model using clinical notes from multiple silos without data transfer. 

2. We show that it is possible to do federated fine tuning of  BERT model for different down stream tasks such as name entity recognition(NER).  

\section{Related Work}

At the beginning of 2019, Lee et al. published a BERT model pre-trained on biomedical PubMed abstracts and PubMedcal central articles~\cite{lee2019biobert}. Later in the year, \citet{alsentzer2019publicly} and \citet{si2019enhancing} published almost at the same time BERT models pre-trained trained on publicly available clinical notes from MIMIC3 either starting from trained parameters of original BERT or BioBERT model and show improvement of clinical NLP tasks. There has been only limited work on federated NLP works in the clincal domain. In 2019, \citet{liu2019two} published a study doing federated training of machine learning models on clinical notes and used the model for patient level phenotyping based on Concept Unique Identifiers (CUIs). In comparison, in out study, we use raw clinical texts instead of CUIs and, to the best of our knowledge, we are the first one to use contextual representation methods for federated clinical NLP tasks.

\begin{figure*}[t]
  \centering 
  \includegraphics[width=4in]{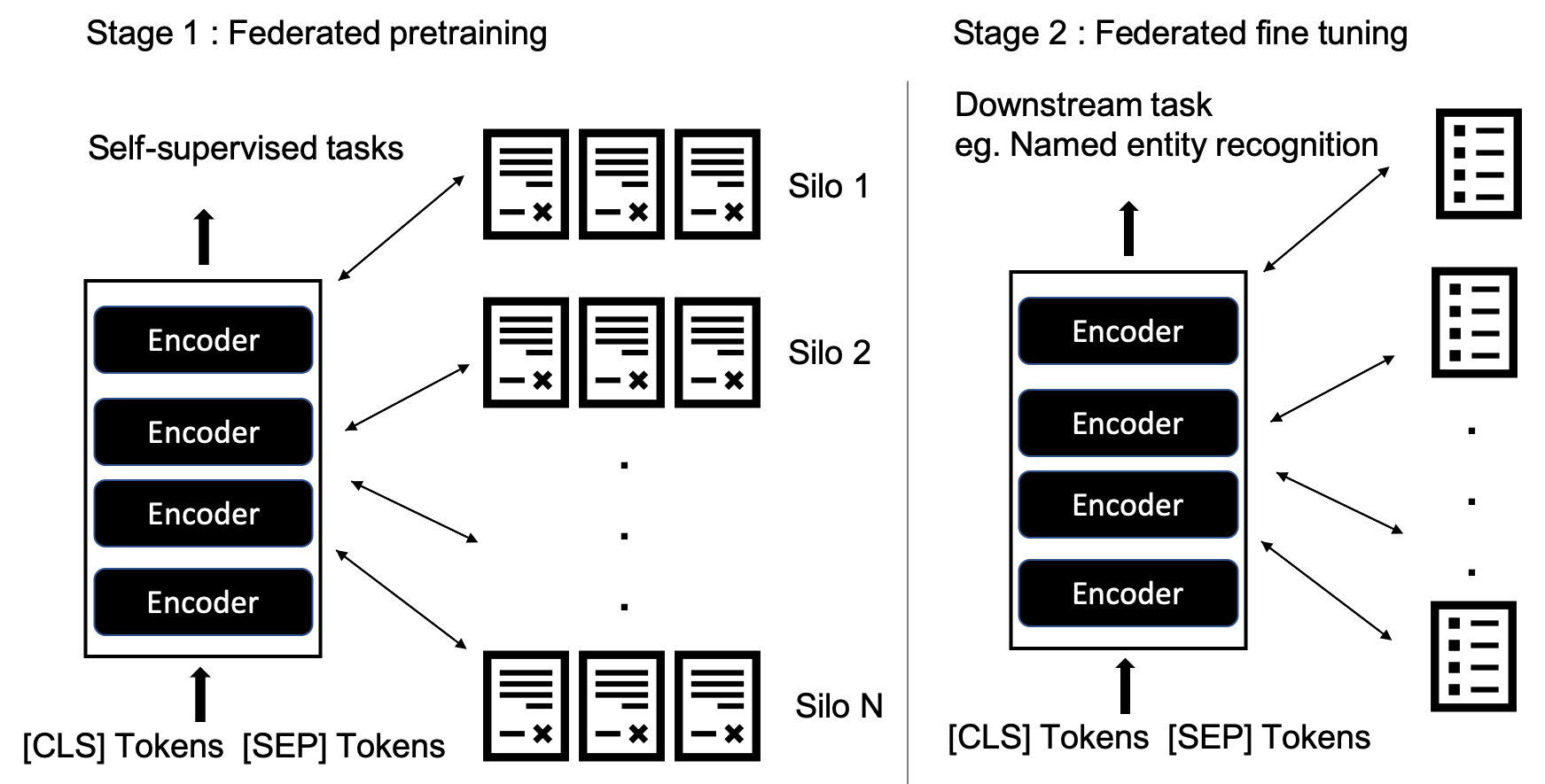} 
  \caption{Federated BERT model trained can be conducted at both pre-training and fine tuning stages. In federated pre-training stage, unlabelled clinical texts from different silos , such as hospitals, are used for self-supervised training to learn domain-specific linguistic characterises in a federated manner with moving data outside their silos. In task-specific fine tuning stage, pre-trained BERT model were further trained using labelled texts from different silos in a federated manner.  }
  \label{fig:Two_stage} 
\end{figure*}

\section{Methods}
\subsection{Clinical data }
 In this study, the publicly available MIMIC-III corpus~\cite{johnson2016mimic} was used for contextual representation learning. This corpus contains information for more than 58,000 admissions for more than 45,000 patients admitted to Beth Israel Deaconess Medical Center in Boston between 2001 and 2012.
 
 Different type of clinical notes are available in this corpus including discharge summaries, nursing notes and so on. We included only discharge summaries in our study as previous studies have shown that performance of a model trained on only discharge summaries in this corpus is only marginally worse than model trained on all notes types\cite{alsentzer2019publicly}. 

\subsection{Federated BERT training }


Our goal is to develop methods for federated learning for both (1) pre-training models to capture linguistic characteristics of clinical text, and (2) fine-tuning for a specific down stream task, here named entity recognition (NER) from clinical notes. These methods will allow researchers and clinicians to utilize data from multiple health care providers to train contextual representation models like BERT, without the need to share the data directly, obviating issues related to data transfer and privacy.

In the following sections, we first describe data processing and a simple notes pre-processing step. We then discuss the method for federated pre-training of BERT model and the method for fine tuning. 

In pre-training sage, clinical notes MIMIC-III corpus were randomly split into 5 groups to mimic 5 different silos by patient. The prepossessing and tokenization pipeline from Alsentzer et al. 2019 \cite{alsentzer2019publicly} was adapted.

To train the BERT model, we simulated sending out models with identical initial parameters to all silos. At each silo, a model was trained using only data from that site. Only model parameters of the models were then sent back to the analyzer for aggregation. An updated model is generated by averaging the parameters of models distributively trained, weighted by sample size \cite{konevcny2016federated,mcmahan2016communication}. In this study, sample size is defined as the number of patients in the pre-training stage and number of notes in fine-tuning stage. 
 
After model aggregation, the updated model was sent out to all sites again to repeat the global training cycle. Formally, the weight update is specified by:
 
\begin{equation}
Q^{t}_{ag}=\sum_{k=1}^{K} \frac{n_k}{N} Q^{t}_{k}
\end{equation}

where $Q^{t}_{ag}$ is the parameter of aggregated model at global cycle $t$, $K$ is the number of data silos. $n_k$ is the number of samples at the $k^{th}$ site, $N$ is the total number of samples across all sites, and $Q_k$ is the parameters learned from the $k^{th}$ data site alone. $t$ is the global cycle number in the range of [1,T]. 

In the pre-training stage, in each global cycle the BERT model was trained for one epoch through all clinical notes data at each of the silos using the default settings from the original BERT publication \cite{devlin2018bert}. A total of 15 global cycles were run. The downstream task performance plateaued out around cycle 10. Therefore, BERT model federately trained for 10 global cycles was used for down stream tasks. Centralized fine-tuning of the i2b2 NER tasks plateaued after 4 epochs, with the learning rate set at $2e-5$ and a batch size of 32 \cite{alsentzer2019publicly}. When conducting federated fine tuning using the same settings as centralized fine tuning, one epoch of training was conducted in each global cycle and a total of 6 global cycles were conducted when the performance plateaued. 

Each pre-training global cycle took around 4 hours on a Tesla K80 GPU which has a single precision GFLOPs of 5591–8736. A single federated fine tuning global took around 20 mins on the same device. 

\subsection{Downstream tasks}
Clinical BERT pre-trained on MIMIC corpus has been reported to have superior performance on NER tasks in Inside-Outside-Beginning (IOB) format \cite{ramshaw1999text} using i2b2 2010 \cite{uzuner20112010} and 2012 \cite{sun2013evaluating} data\cite{alsentzer2019publicly}.Original training/development/test splits in the challenges were used. 
The NER tasks classify if a token is within a span of a class, outside spans of any classes or at beginning of a span of a class. For example, the sentence "He has severe asthma" is labelled as "Null, Null,B-problem and I-problem". This means the first two words are not in any classes, the third word is the beginning of span of class "problem" and the forth word is in the class "problem".
\par
In this study we use these two data sets for the NER tasks and for fine tuning.To conduct federated fine tuning of BERT model, training notes for downstream task were randomly split into 5 silos by note. 


\section{Experiments and results}

\subsection{Experimental design}
In order to understand whether large size contextual language representation model like BERT can be pre-trained and fine tuned in a federated manner using data from different silos, we designed and conducted the following 6 experiments. 

First of all, we looked at the scenarios no domain specific BERT model pre-training was conducted. In those cases, the parameters (checkpoint) from original BERT base model trained on Books Corpus  and English Wikipedia \cite{zhu2015aligning,devlin2018bert}. We also looked at the scenarios where BERTbase model was pre-trained by MIMIC3 discharge summaries in a centralized manner\cite{alsentzer2019publicly}. Lastly, we look at the scenarios where BERTbase model was pre-trained by MIMIC3 discharge summaries in a federated manner. For each of these conditions, we fine tuned BERT models for downstream tasks using centralized vs. federated learning.

To summarize, six experiments were conducted: 

\begin{enumerate}
\itemsep0em
\item Original BERT + centralized fine tuning
\item Clinical BERT + centralized fine tuning
\item Federated clinical BERT + centralized fine tuning
\item Original BERT + federated fine tuning 
\item Clinical BERT + federated fine tuning 
\item Federated clinical BERT + federated fine tuning.
\end{enumerate}

\begin{table}[t!]
\centering 

  \caption{Performance on i2b2 NER tasks}
\resizebox{\columnwidth}{!}{%
\begin{tabular}{l|l|l|l|l|l}
\textbf{Task} & \textbf{Pretraining} & \textbf{fine tuning} & \textbf{Prec} & \textbf{Rec} & \textbf{F1} \\ \hline
i2b2\_2010        & BERTbase                  & Centralized                 & 0.775              & 0.794           & 0.784       \\ \hline
i2b2\_2010        & ClinicalBERT              & Centralized                 & 0.844              & 0.873           & 0.858       \\ \hline
i2b2\_2010        & Fed\_ClinicalBERT         & Centralized                 & 0.81               & 0.831           & 0.820        \\ \hline
i2b2\_2010        & BERTbase                  & Federated                   & 0.73               & 0.703           & 0.716       \\ \hline
i2b2\_2010        & ClinicalBERT              & Federated                   & 0.819              & 0.868           & 0.843       \\ \hline
 i2b2\_2010        & Fed\_ClinicalBERT         & Federated                   & 0.811              & 0.806           & 0.808       \\ \hline
\hline
i2b2\_2012        & BERTbase                  & Centralized                 & 0.704              & 0.754           & 0.728       \\ \hline
i2b2\_2012        & ClinicalBERT              & Centralized                 & 0.711              & 0.774           & 0.741       \\ \hline
i2b2\_2012        & Fed\_ClinicalBERT         & Centralized                 & 0.708              & 0.764           & 0.735       \\ \hline
i2b2\_2012        & BERTbase                  & Federated                   & 0.667              & 0.707           & 0.686       \\ \hline
i2b2\_2012        & ClinicalBERT              & Federated                   & 0.697              & 0.769           & 0.731       \\ \hline
i2b2\_2012        & Fed\_ClinicalBERT         & Federated                   & 0.687              & 0.745           & 0.715       
\end{tabular}
}
\label{Table_results}
\end{table}

\subsection{Experimental results}

The results of our experiments are shown in Table~\ref{Table_results}. In experiment 1, when original BERT was used without domain-specific pre-training on MIMIC discharge summary and fine tuning was conducted in a centralized manner, the F1 score of 0.784 was achieved for i2b2 2010 and 0.728 for i2b2 2012. In experiment 2, where both BERT pre-training and fine tuning was conducted in a centralized manner, the F1 was 0.858 for i2b2 2010 and 0.741 for i2b2 2012. In experiment 3, where BERT was pre-trained in a federated manner and fine tuned using centralized data, the F1 was 0.820 for i2b2 2010 and 0.735 for i2b2 2012.

In Experiment 4 where original BERT was not pre-trained using  MIMIC3 discharge summary and fine tuning was conducted in a federated manner, the F1 was 0.716 for i2b2 2010 and 0.686 for i2b2 2012. In comparison, if the BERT is trained using centralized clinical notes before federated fine tuning (Experiment 5), F1 scores of i2b2 2010 NER task improved to 0.843 and  F1 scores of i2b2 2012 NER task improved to 0.731.  While the both the pre-training and fine tuning were conducted in a federated, as in Experiment 6, the F1 score were 0.808 and 0.715 for i2b2 2010 and i2b2 2012 respectively, which is superior to BERT model without pre-training. We made our BERT model federatedly trained with discharge summaries publicly available at XXXX and all the codes at XXXX. 

\subsection{Attention analysis}
To understand how different BERT model works, we look at behaviors of attention head. Firstly, entropy level of a each head's attention distribution was analyzed\cite{clark2019does}.We compared original BERTbase model, Clinical BERT model and federatedly trained clinical BERT model without fine tuning in this analysis. Attention heads from all three models show decreasing entropy with depth of layers.  By comparing correlation of attention entropy among heads in different model, we found the federated clinical BERT model is more similar to BERTbase model than centralized clinical BERT model(table 2).
\par 

We analyzes the attention from from three different types of BERTs models to understand if if there are different clustering patterns. We define distance among Jensen-Shannon Divergence matrices of attentions heads from different models as mean of absolute value of element-wide difference. The models pre-trained with clinical notes are similar to each other, while BERT base model have a large distance with other two models.

\begin{table}[t!]
\centering
\caption{Analysis of attention heads}
\resizebox{\columnwidth}{!}{%
\begin{tabular}{llll}
\multicolumn{4}{c}{\textbf{Spearman correlation among entropy of attention heads}}                                                                   \\ \hline
\multicolumn{1}{|l|}{}                  & \multicolumn{1}{l|}{BERTbase} & \multicolumn{1}{l|}{ClinicalBERT} & \multicolumn{1}{l|}{Fed\_ClinicalBERT} \\ \hline
\multicolumn{1}{|l|}{BERTbase}          & \multicolumn{1}{l|}{1}        & \multicolumn{1}{l|}{}             & \multicolumn{1}{l|}{}                  \\ \hline
\multicolumn{1}{|l|}{ClinicalBERT}      & \multicolumn{1}{l|}{0.25}     & \multicolumn{1}{l|}{1}            & \multicolumn{1}{l|}{}                  \\ \hline
\multicolumn{1}{|l|}{Fed\_ClinicalBERT} & \multicolumn{1}{l|}{0.96}     & \multicolumn{1}{l|}{0.27}         & \multicolumn{1}{l|}{1}                 \\ \hline
\multicolumn{4}{c}{\textbf{Distances among Jensen-Shannon Divergence matrices of attention heads}}                                                   \\ \hline
\multicolumn{1}{|l|}{}                  & \multicolumn{1}{l|}{BERTbase} & \multicolumn{1}{l|}{ClinicalBERT} & \multicolumn{1}{l|}{Fed\_ClinicalBERT} \\ \hline
\multicolumn{1}{|l|}{BERTbase}          & \multicolumn{1}{l|}{0}        & \multicolumn{1}{l|}{}             & \multicolumn{1}{l|}{}                  \\ \hline
\multicolumn{1}{|l|}{ClinicalBERT}      & \multicolumn{1}{l|}{8398.95}  & \multicolumn{1}{l|}{0}            & \multicolumn{1}{l|}{}                  \\ \hline
\multicolumn{1}{|l|}{Fed\_ClinicalBERT} & \multicolumn{1}{l|}{8153.26}  & \multicolumn{1}{l|}{314.38}       & \multicolumn{1}{l|}{0}                 \\ \hline
\end{tabular}
}
\end{table}

\section{Discussion and conclusion}

In this proof-of-concept study, we demonstrate the possibility to both pre-train and fine tune large contextual representation language model BERT in a federated manner.

\par 
Our analyses suggest that conducting pre-training and fine tuning in a federated manner using data from different silos resulted in reduced performance compared with training on centralized data. We can refer to this loss of performance due to separation of data as "federated communication loss". When only conducting pre-training on discharge summaries in a federated manner and fine tuning model for NER tasks in a centralized manner, performance on downstream tasks only had a less than $5\%$ drop compared with model both pre-trained and fine-tuned in a centralized manner. When only conducting fine tuning in a federated manner but conduct pre-training kept with centralized data, the performance dropped less than 2\%. However, when conducting both pre-training and fine tuning in a federate manner, the performance had a non negligible decrease of around 6\%.  

\par 

\par
 There are several limitations in this study. Firstly of all, due to limit of data access, we used clinical notes from a single healthcare system to simulate different silos. 
In future studies ,we would like to conduct analysis on clinical notes from a wide range of health care providers. Secondly, the MIMIC3 training settings has a certain percentage of overlaps with the tasks text in i2b2, which could cause confusion in model performance. This is the reason why we did not include MedNLI tasks in out study, as all of MedNLI texts are directly from MIMIC3 database.In future study, more data-sets without any overlapped data should be used.





\bibliography{anthology,acl2020}
\bibliographystyle{acl_natbib}

\end{document}